\documentclass{bmvc2k}

\usepackage{graphicx}
\usepackage{amsmath}
\usepackage{amssymb}
\usepackage{booktabs}

\usepackage{graphicx}
\usepackage{amsmath}
\usepackage{amssymb}
\usepackage{booktabs}

\usepackage{amsmath,amsfonts,bm, bbm}

\def\eqref#1{equation~\ref{#1}}

\def\1{\bm{1}}

\def\va{{\bm{a}}}

\DeclareMathAlphabet{\mathsfit}{\encodingdefault}{\sfdefault}{m}{sl}
\SetMathAlphabet{\mathsfit}{bold}{\encodingdefault}{\sfdefault}{bx}{n}

\def\gA{{\mathcal{A}}}
\def\gB{{\mathcal{B}}}
\def\gC{{\mathcal{C}}}
\def\gD{{\mathcal{D}}}

\def\gF{{\mathcal{F}}}

\def\gH{{\mathcal{H}}}

\def\gX{{\mathcal{X}}}

\def\gZ{{\mathcal{Z}}}

\newcommand{\hx}{\hat{x}}

\newcommand{\ft}{f_\theta}

\newcommand{\norm}[1]{\lVert #1 \rVert}

\newcommand{\bi}{\mathbbm{1}} 
\usepackage[utf8]{inputenc} 
\usepackage[T1]{fontenc}    
\usepackage{url}            
\usepackage{booktabs}       
\usepackage{amsfonts}       
\usepackage{nicefrac}       
\usepackage{microtype}      
\usepackage{xcolor}         
\usepackage{multirow}
\usepackage{dsfont}
\usepackage{soul}
\usepackage{adjustbox}
\usepackage{xspace}
\usepackage{bm}
\usepackage{amsmath}
\usepackage{setspace}                       
\usepackage{scrextend}
\usepackage{mathtools}
\usepackage{graphicx}
\usepackage{subcaption}
\usepackage{enumitem}
\usepackage{colortbl}
\usepackage{wrapfig}
\usepackage{kotex}
\usepackage{lipsum}
\usepackage{xcolor,pifont}
\usepackage{algpseudocode,algorithm,algorithmicx,}
\usepackage{tikz}

\definecolor{myyellow}{rgb}{1,1, 0.6}
\definecolor{myorange}{rgb}{1, 0.8, 0.6}
\definecolor{myred}{rgb}{1, 0.6, 0.6}

\title{Attribute-Guided Encryption with Facial Texture Masking}

\addauthor{Chun Pong Lau}{clau13@jhu.edu}{1}
\addauthor{Jiang Liu}{jiangliu@jhu.edu}{1}
\addauthor{Rama Chellappa}{rchella4@jhu.edu}{1}

\addinstitution{
 Johns Hopkins University\\
 Baltimore, USA
}

\runninghead{Lau et al.}{Attribute Guided Encryption with Facial Texture Masking}

\begin{document}

\maketitle

\begin{abstract}
The increasingly pervasive facial recognition (FR) systems raise serious concerns about personal privacy, especially for billions of users who have publicly shared their photos on social media. Several attempts have been made to protect individuals from unauthorized FR systems utilizing adversarial attacks to generate encrypted face images to protect users from being identified by FR systems. However, existing methods suffer from poor visual quality or low attack success rates, which limit their usability in practice. In this paper, we propose Attribute Guided Encryption with Facial Texture Masking (AGE-FTM) that performs a \textit{dual manifold} adversarial attack on FR systems to achieve both good visual quality and high black-box attack success rates. In particular, AGE-FTM utilizes a high fidelity generative adversarial network (GAN) to generate natural \textit{on-manifold} adversarial samples by modifying facial attributes, and performs the facial texture masking attack to generate imperceptible \textit{off-manifold} adversarial samples. Extensive experiments on the CelebA-HQ dataset demonstrate that our proposed method produces more natural-looking encrypted images than state-of-the-art methods while achieving competitive attack performance. We further evaluate the effectiveness of AGE-FTM in the real world using a commercial FR API and validate its usefulness in practice through an user study.
\end{abstract}
\setlength{\abovedisplayskip}{3pt}
\setlength{\belowdisplayskip}{3pt}
\setlength{\abovecaptionskip}{5pt}
\setlength{\textfloatsep}{5pt}
\setlength{\parskip}{-2pt plus2pt minus2pt}


\section{Introduction} \label{sec:Introduction}
\looseness -1 The rise of deep neural networks has enabled the tremendous success of facial recognition (FR) systems~\cite{deng2019arcface, schroff2015facenet, mobileface}. However, the widely deployed FR systems also pose significant threat to personal privacy as billions of users have publicly shared their photos on social media. By large-scale social media photo analysis, FR systems can be used for detecting user relationships~\cite{10.1515/popets-2015-0004},  stalking victims~\cite{shwayder2020clearview}, stealing identities~\cite{lively}, and performing massive government surveillance~\cite{satariano2019police, hill2020secretive, mozur2019surveillance}. It is critical to develop facial privacy protection techniques to protect individuals from unauthorized FR systems.
\begin{figure}[t]
\centering
\setlength\tabcolsep{1pt}
\scalebox{1.0}{\begin{tabular}{cccc}
\begin{subfigure}[t]{0.24\textwidth}
\includegraphics[width=\textwidth]{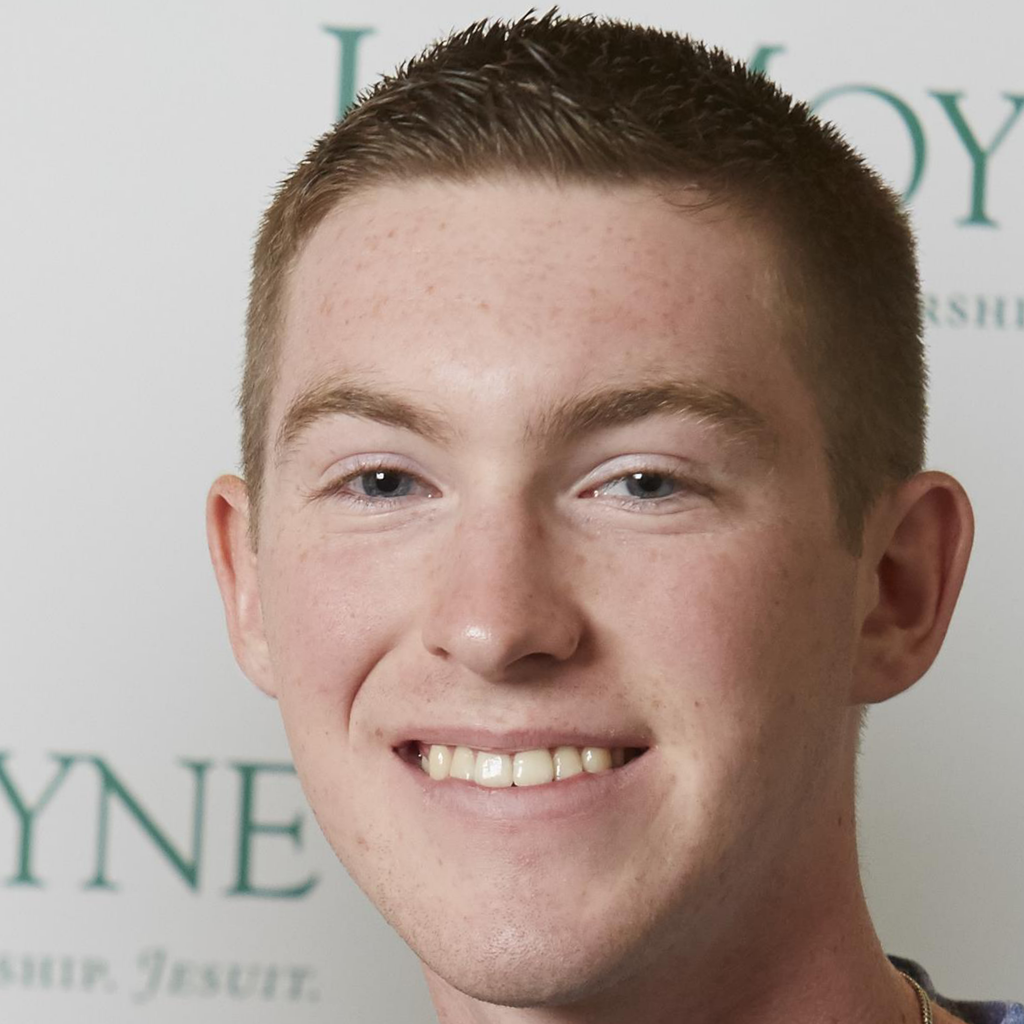}
\vspace{-11mm}
\begin{flushleft}
\small{\textcolor{blue}{\textbf{46.98}}}
\end{flushleft}
\vspace{-2mm}
\caption{Original}
\end{subfigure} 
&
\begin{subfigure}[t]{0.24\textwidth}
\includegraphics[width=\textwidth]{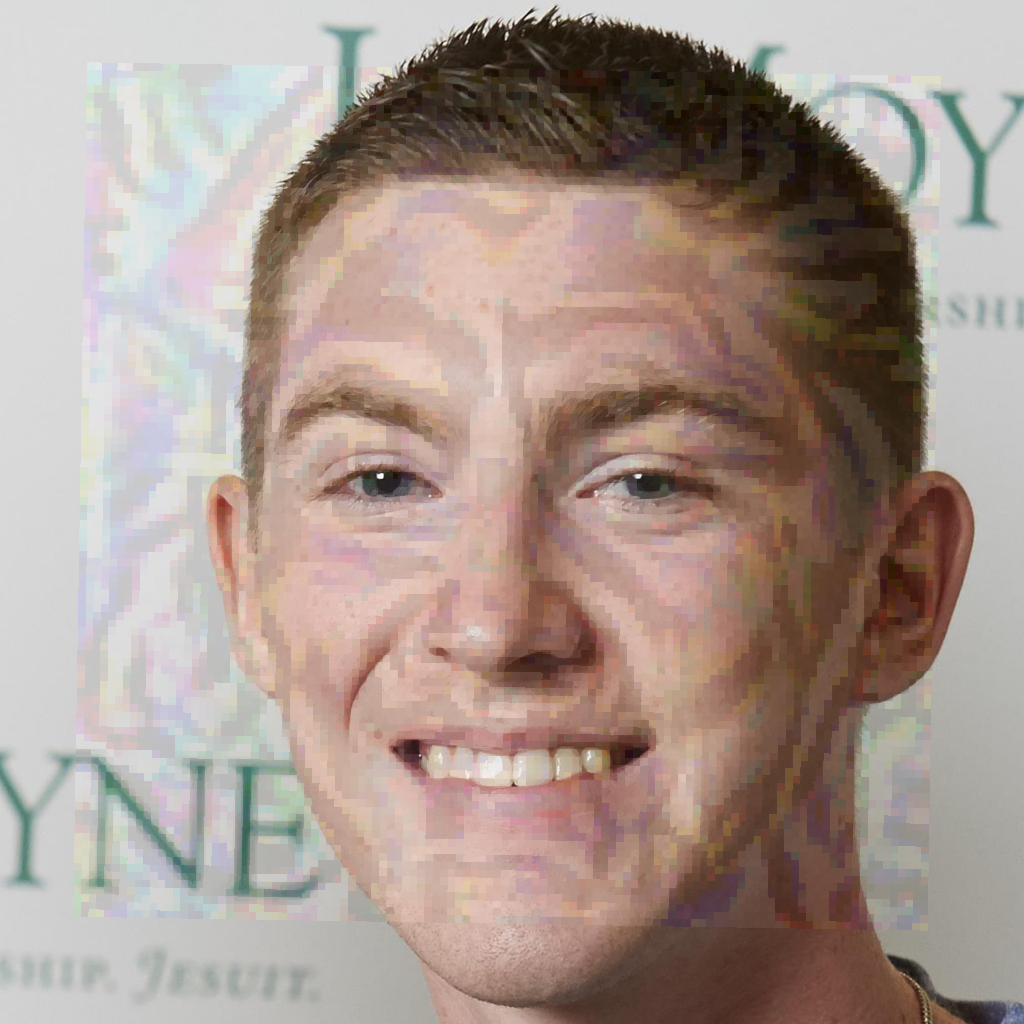}
\vspace{-11mm}
\begin{flushleft}
\small{\textcolor{blue}{\textbf{64.56}}}
\end{flushleft}
\vspace{-2mm}
\caption{TIP-IM \cite{yang2021towards} }
\end{subfigure} &
\begin{subfigure}[t]{0.24\textwidth}
\includegraphics[width=\textwidth]{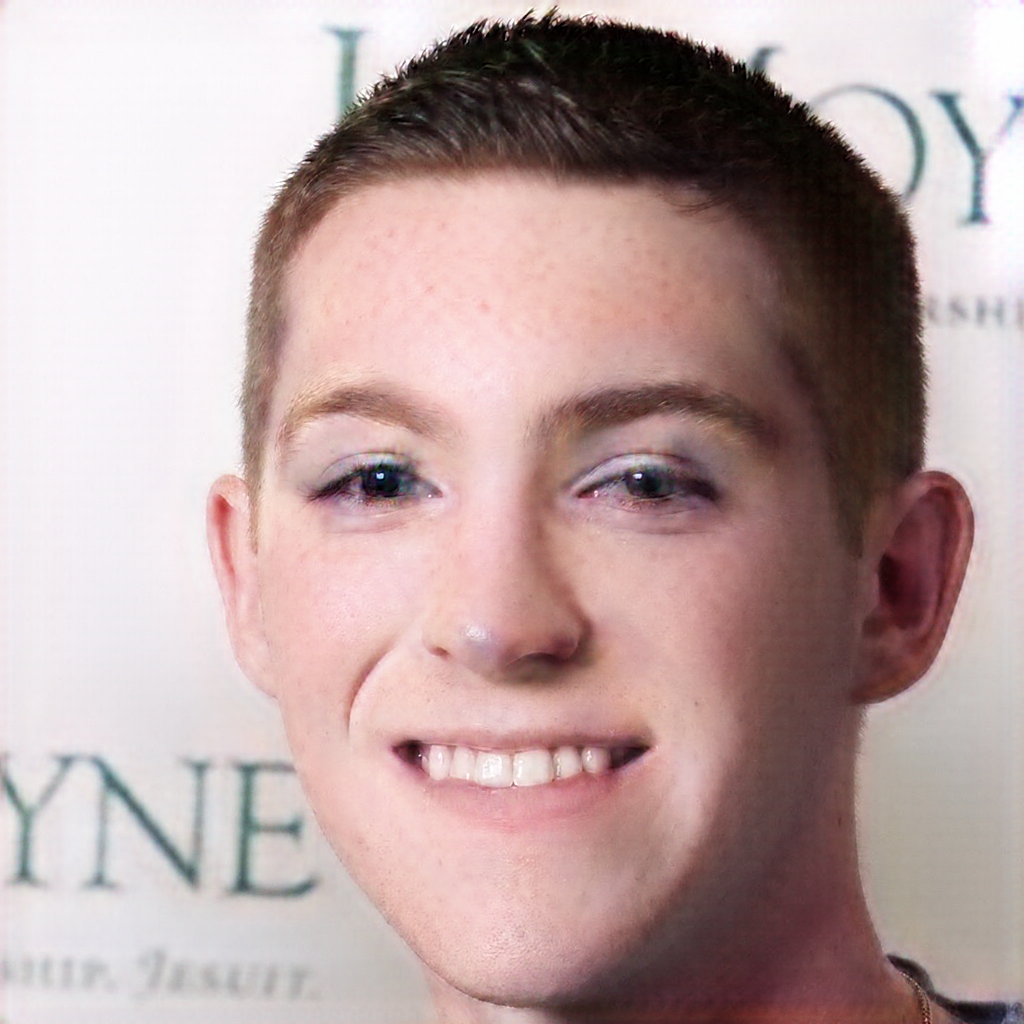}
\vspace{-11mm}
\begin{flushleft}
\small{\textcolor{blue}{\textbf{49.93}}}
\end{flushleft}
\vspace{-2mm}
\caption{AMT-GAN \cite{hu2022protecting} }
\end{subfigure} &
\begin{subfigure}[t]{0.24\textwidth}
\includegraphics[width=\textwidth]{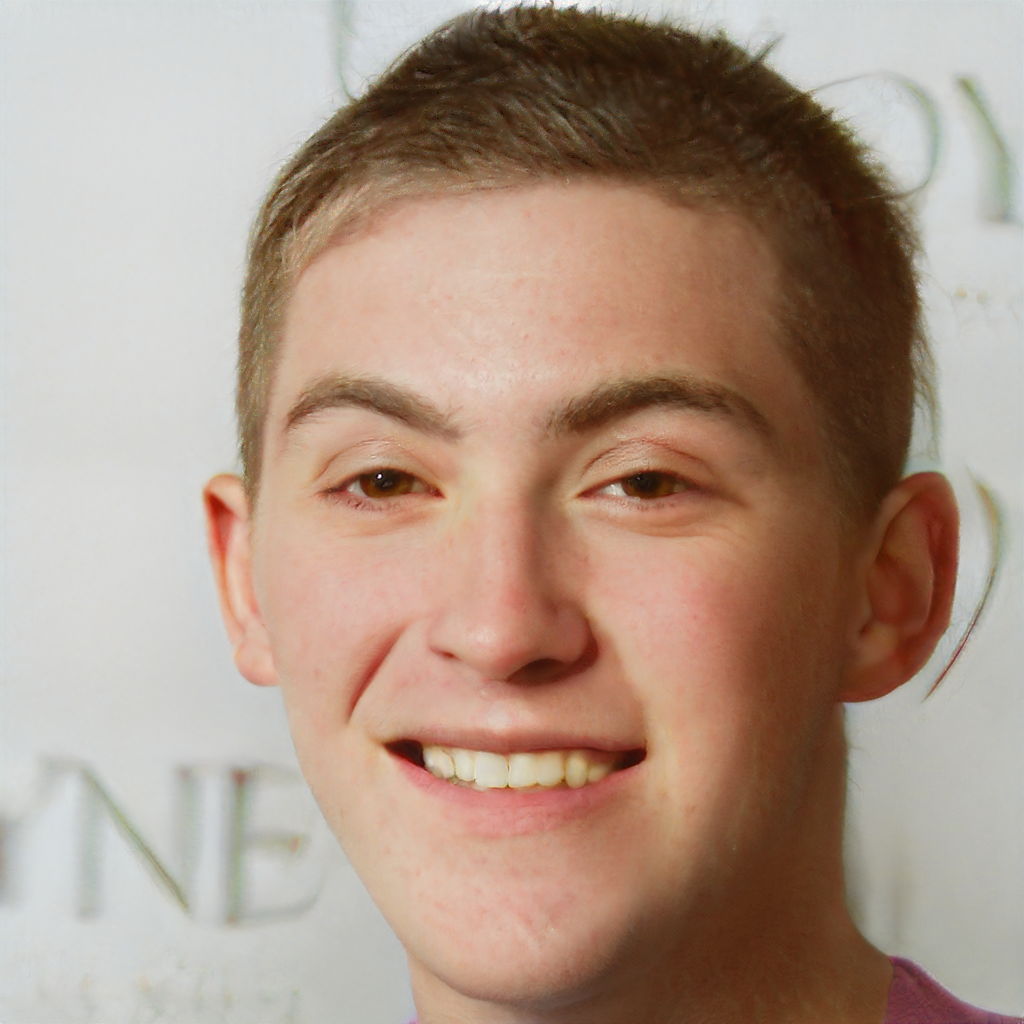}
\vspace{-11mm}
\begin{flushleft}
\small{\textcolor{blue}{\textbf{63.49}}}
\end{flushleft}
\vspace{-2mm}
\caption{AGE-FTM (ours) }
\end{subfigure} 
\end{tabular}}

\caption{{Comparison with other black-box adversarial attacks on FR systems. The number on the left bottom corner is the face verification confidence score with a target image of other person from a commercial FR API Face++.}}
\label{fig:teaser}
\end{figure}

Recently, several works~\cite{cherepanova2021lowkey, hu2022protecting} used adversarial attacks to generate encrypted face images to protect users from being identified by FR systems. However, existing attacks on FR systems~{\cite{cherepanova2021lowkey, hu2022protecting, komkov2021advhat, sharif2019general, yin2021adv, yang2021towards, zhong2020towards}} suffer from poor visual quality or low attack success rates. Noise-based methods such as Lowkey~\cite{cherepanova2021lowkey} and TIP-IM~\cite{yang2021towards} create unpleasant noise patterns on face images(see Fig. ~\ref{fig:teaser}(b)). Patch attack-based methods such as Adv-Hat~\cite{komkov2021advhat} and Adv-Glasses~\cite{sharif2019general} add unexplaniable and conspicuous changes to the source images. Makeup-based methods~\cite{yin2021adv, hu2022protecting} drastically change the makeup styles of source images and might be biased towards female users~\cite{hu2022protecting} (see Fig. ~\ref{fig:teaser}(c)). From a user-centered perspective, the aforementioned approaches are not desirable in practice, as everyone wants to post their best-looking photos on social media -- not unattractive photos, even though they might be encrypted for protecting personal privacy. 

\begin{figure*}[t]
\centering
\includegraphics[width=\textwidth]{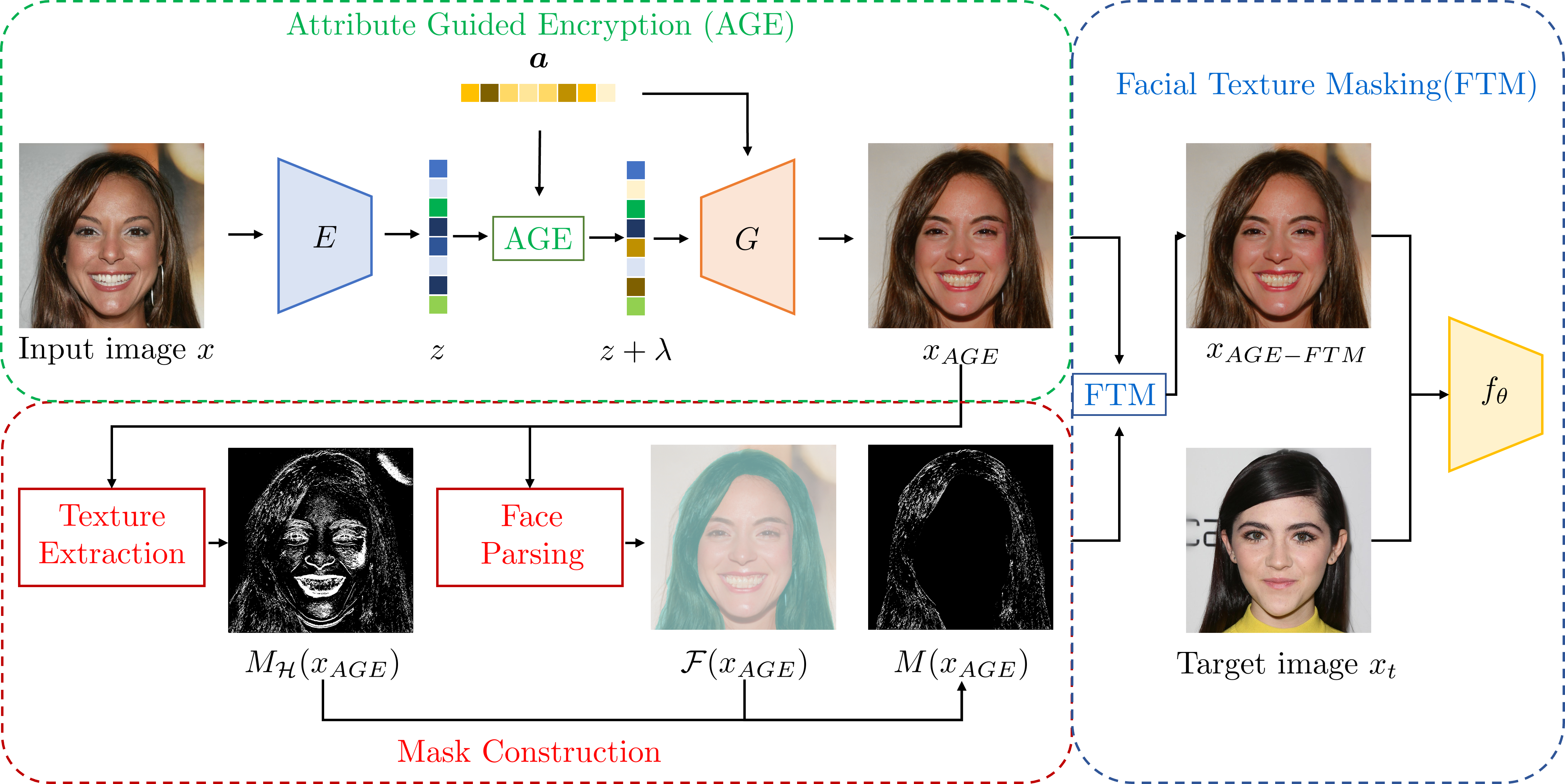}

\caption{Overview of AGE-FTM. The face image $x$ is passed through the AGE, which is an on-manifold attack. The corresponding latent code $z$, which is encoded by $E$, is perturbed within the latent space and guided by the attribute vector. Then the intermediate image $x_{AGE}$, is passed through the texture extraction module and the face parsing module to obtain the texture mask and face parsing map respectively. These two masks will be combined as a hair texture mask $M$. Then $x_{AGE}$ will be passed through FTM, which is a off-manifold attack restricted on the mask $M$, to obtain the resultant protected image $x_{AGE-FTM}$ }
\label{fig: overview}
\end{figure*}

An ideal facial identity encryption algorithm should only create \textit{natural} or \textit{imperceptible} changes to the source images, while being able to fool the FR systems. To achieve this goal, we propose Attribute Guided Encryption with Facial Texture Masking (AGE-FTM) that performs dual manifold adversarial attacks on FR systems. We utilize a high fidelity generative adversarial network (GAN) to learn the natural face manifold, and propose the Attributed Guided Encryption (AGE) strategy that uses the attribute vectors to guide the \textit{on-manifold} attacks, which ensures the encrypted images are \textit{natural-looking}. {Unlike makeup-based methods that induce severe artifacts on male faces image, changing attribute vectors such as "age" and "smile" does not have significant gender bias and hence both the protected male and female images look naturally.} In addition, AGE-FTM performs Facial Texture Masking (FTM) Attack that generates \textit{off-manifold} adversarial samples by only perturbing the texture regions of the images. Human perception systems are less sensitive to changes in high-frequency components compared to low-frequency components~\cite{hvs}, which ensures the perturbations generated by FTM to be \textit{imperceptible}. The overview of AGE-FTM is shown in Fig.~\ref{fig: overview}.

In summary, the contributions of this paper are as follows: 
\begin{itemize}
    \item We propose AGE-FTM, which performs dual manifold adversarial attacks on FR systems. The off-manifold attack FTM only perturbs pixels in the hair texture region. On the other hand, the on-manifold attack AGE applies attribute vectors to guide the latent space attack. These two components ensure AGE-FTM to create natural and imperceptible changes for facial identity protection.
    \item Our extensive experiments on the CelebA-HQ dataset demonstrate that AGE-FTM produces more natural looking encrypted images than state-of-the-art methods while achieving competitive attack performance. 
    \item We further evaluate the effectiveness of AGE-FTM in the real world using a commercial FR API Face++\footnote{\url{https://www.faceplusplus.com/}} and validate its usefulness in practice through a user study.
\end{itemize}

\section{Related Work}
\subsection{Adversarial Attacks on Face Recognition}
Many studies have been proposed to attack FR systems, including both poisoning~\cite{shan2020fawkes} and evasion~\cite{cherepanova2021lowkey, hu2022protecting, komkov2021advhat, sharif2019general, yin2021adv, yang2021towards} attacks. Poisoning attacks require injecting poisoned face images into the training sets of FR systems, which is unlikely to achieve for individual users. Evasion attacks, especially transferable black-box attacks, are more practical for protecting facial image privacy, as they only require perturbing the source images to fool the FR systems. Existing attacks on FR systems~\cite{cherepanova2021lowkey, hu2022protecting, komkov2021advhat, sharif2019general, yin2021adv, yang2021towards} suffer from poor visual quality or low attack success rates, which limit their usability in the real world. Noise-based methods such as Lowkey~\cite{cherepanova2021lowkey} and TIP-IM~\cite{yang2021towards} create unexplainable noise patterns on face images. Patch attack-based methods such as Adv-Hat~\cite{komkov2021advhat} and Adv-Glasses~\cite{sharif2019general} add unnatural and conspicuous changes to the source images. Makeup based-methods~\cite{yin2021adv, hu2022protecting} drastically change the makeup styles of source images and might be biased towards female users~\cite{hu2022protecting}. In contrast, the proposed AGE-FTM creates natural and imperceptible changes to the source images and achieves competitive attack performance.

\subsection{Off- and On-Manifold Attacks}
\looseness -1 Recently, several works aim to use a low-dimensional underlying data manifold to attack neural networks by creating on-manifold adversarial samples \cite{jalal2017robust,song2018adv,Stutz_2019_CVPR, lin2020dual}. On-manifold adversarial samples are adversarial samples constrained to lie on data manifolds and are obtained by perturbing inputs in the latent space learned by generative models. Adversarial samples computed in the image space are considered as off-manifold \cite{Stutz_2019_CVPR}. On-manifold adversarial samples have been used to break models trained by adversarial training \cite{song2018adv} as well as deep generative model-based defense methods \cite{Chen2020onbreaking} such as DefenseGAN \cite{samangouei2018defense}, Analysis by Synthetics \cite{schott2018towards} and MoG-VAE \cite{ghosh2019resisting}. Compared to off-manifold attacks, on-manifold attacks generate more natural-looking and less conspicuous adversarial samples as they come from a learned natural image manifold. \cite{lau2021ijsat} proposed Joint Space Threat Model, which is a dual-manifold attack that combines both on-manifold and off-manifold perturbations.
\section{Attribute Guided Encryption with Facial Texture Masking (AGE-FTM)}
\subsection{Problem Formulation}
In this section, we formulate the problem of adversarial attacks on FR systems. Suppose the face images $x\in \gX := \mathbb{R}^{H \times W \times C}$ are drawn from an underlying distribution $\mathbb{P}_X$, where $H$, $W$ and $C$ are the height, width and the number of channels of the image respectively. Let $f_\theta$ be a parameterized model which maps any image in $\gX$ to a feature vector $y$ in $f_\theta(\gX)$. An \emph{accurate} FR system can map two images $x_1$ and $x_2$ with the same identity to features that are very close in the feature space, i.e. $\gD(\ft(x_1), \ft(x_2)) \leq \tau$, where $\gD$ is a distance function and $\tau$ is a threshold. A successful attack fools the FR system to map an adversarial image $\hx$ with the same identity as $x$ to a feature that is far away from the feature of $x$, i.e. $\gD(\ft(\hx), \ft(x)) \geq \tau$. Another type of successful attack is the targeted attack, which is also called the impersonation attack in FR. It aims to perturb the source image such that it is identified as the target identity, i.e. $\gD(\ft(\hx), \ft(x_t)) \leq \tau$ where $x_t$ is the target image. From \cite{lin2020dual}, \emph{exact manifold assumption (EMA)} is that there exists a generative model $G$ such that there exists a latent representation $z$ in the latent space $\gZ := \mathbb{R}^d$ for every image $x$. 

\paragraph{Off- and On-Manifold Robustness} To encrypt the facial information against malicious FR models in a practical scenario, we consider the impersonation attack in a black-box setting. We use multiple FR models to craft the attack and denote the set of FR systems as $\gA_f$. Formally, we aim to solve the following problem: \begin{equation}
    \min_{\hx} \gD(\ft(\hx), \ft(x_t)) = \sum_{\ft \in \gA_f} 1 - \cos{(\ft(\hx), \ft(x_t))},
    \label{eq:standard}
\end{equation} 
where $x_t$ is the target image and $\gD$ is the cosine similarity loss. 
We consider the adversarial samples perturbed within the image space and the latent space as off-manifold and on-manifold adversarial samples respectively. Mathematically, 
\begin{equation}
    \min_{\delta \in \Delta} \gD(f_{\theta}(x + \delta), f_{\theta}(x_t))~\text{and}~\min_{\lambda \in \Lambda} \gD (f_{\theta}(G(z + \lambda)), f_{\theta}(x_t)),
    \label{eq:off-manifold}
\end{equation}

where $x + \delta$ and $G(z + \lambda)$ are the off-manifold and on-manifold adversarial samples respectively, $\Delta = \{ \delta: \norm{\delta}_p < \epsilon\}$ and $\Lambda = \{ \lambda: \norm{\lambda}_p < \eta\}$.
\subsection{Facial Texture Masking Attack}
Off-manifold attacks in the image space such as PGD \cite{madry2017towards} are usually strong and imperceptible. However, when it comes to FR, in particular for black-box attacks, it becomes less effective. Moreover, it is easier to perceive the existence of the attack when it comes to face images than in other images. According to \cite{hvs}, human perception systems are less sensitive to changes in high-frequency components compared to low-frequency components. In other words, edges or textures of the face image are less disturbing to the human perception systems. On the other hand, faces of people tend to be smooth piecewise regions. It is more noticeable when there is a perturbation in those regions. This motivates us to propose Facial Texture Masking (FTM) Attack, which consists of two components. 

The first one is that we do not attack the whole image but only the textured regions in the image. This makes the attack hard to notice from the perspective of the human visual system. There are many methods to extract the textured regions in the images, such as edge detection algorithms or truncation in the frequency domain using Fourier transform or Wavelet transform. To reduce computational cost, we are intentionally not using a deep learning approach to extract textures. Instead, we apply \emph{unsharp masking} technique to extract the textures. Suppose $\gB$ is a Gaussian blur operator. We can obtain the binary mask $M_\gH$ by 
\begin{equation}
    M_\gH = \bi\{ \gH(x) > \gamma\},~\gH(x) = |x - \gB(x)|
\end{equation}

where $\bi$ is the characteristic function, $\gH$ is the high frequency component that generally consists of edges, and $\gamma$ is some predefined threshold. Then we can restrict the off-manifold perturbation within the high frequency components. Taking the PGD attack as an example, we only perturb the pixels that are in the high frequency regions, i.e. $\gA_\gH = \{ x_{ij} : M_\gH(x_{ij})=1\}$. The iteration steps become:

\begin{equation}
    \delta_{k+1} = \epsilon_{iter} \cdot sign \left(\nabla_{\delta_k|\gA_\gH} \gD(f_{\theta}(x + \delta_k), f_{\theta}(x_t)) \right),~ \nabla_{\delta_k|\gA_\gH}(x) = \nabla_{\delta_k}(x) \odot M_\gH,
\end{equation}


\noindent where $\epsilon_{iter}$ is the attack step size at each iteration and $\odot$ is the Hadamard product. We denote this attack as Texture Masking Attack (TMA). 

Although the TMA significantly reduces the perceptibility of the attack, we can still notice the perturbations at the face portion if we zoom in and look carefully. To further improve the imperceptibility of the attack, we observe that perturbations within the hair region are barely noticable. Therefore, we leverage a face parsing algorithm to locate the hair region. Suppose $\gF$ be the face parsing algorithm, which is a function $\gF: \gX \rightarrow \gC$ that maps each pixel $x_{ij}$ in a face image $x$ in $\gX$ to a discrete component label $c$ in $\gC:= \{1, \cdots, |\gC| \}$. Then the new binary mask $M$ is
\begin{equation}
    M = \bi\left\{ \{\gH(x) > \gamma\} \cap \{x_{ij}: \gF(x_{ij}) = c_{\text{hair}} \} \right\},
\end{equation}
\noindent where $\{x_{ij}: \gF(x_{ij}) = c_{\text{hair}} \}$ is the set containing the pixels in the hair portion. Denote $\gA = \{ x_{ij} : M(x_{ij})=1\}$. The formulation of the FTM attack becomes
\begin{equation}
    \min_{\delta \in \Delta} \gD(f_{\theta}(x + \delta \odot M), f_{\theta}(x_t)),
    \label{eq:FTM}
\end{equation}

\noindent and the iteration steps in the FTM attack will be 
\begin{equation}
    \delta_{k+1} = \epsilon_{iter} \cdot sign \left(\nabla_{\delta_k|\gA} \gD(f_{\theta}(x + \delta_k), f_{\theta}(x_t)) \right).
\end{equation}

\subsection{Attributed Guided Encryption}
Makeup attacks such as \cite{yin2021adv,hu2022protecting} are shown to be effective for FR systems. These attacks can synthesize makeup in some parts of the face, such as the eye region in  \cite{yin2021adv} or the whole face in \cite{hu2022protecting}. However, there are some drawbacks. Makeup attacks usually require a makeup dataset for training. This leads to better visual performance for female because of the imbalance of gender in the makeup dataset. Moreover, in order to increase the attack strength, makeup attacks usually produce heavy makeup on the original images which may be undesirable. Therefore, we propose the Attributed Guided Encryption (AGE), which uses the attribute vectors in the generator to guide the on-manifold attack. We apply HFGI \cite{wang2021HFGI}, which is a state-of-the-art method that uses a high fidelity GAN and can invert and edit high resolution face images effectively. HFGI consists of an encoder $E$ to encode the images and uses a generator $G$ to edit the high resolution face images. Then the AGE formulation is 
\begin{equation}
    \min_{\lambda \in \Lambda} \gD (f_{\theta}(G(E(x) + \lambda;\va )), f_{\theta}(x_t)),
    \label{eq:AGE}
\end{equation}


\noindent To obtain the perturbation, we optimize it by
\begin{equation}
    \lambda_{k+1} = \eta_{iter} \cdot sign \left(\nabla_{\lambda_k} \gD(f_{\theta}(G(E(x) + \lambda_k;\va_k), f_{\theta}(x_t)) \right),
\end{equation}
\noindent where $\va_k = \dfrac{k\va}{N}$,  $\eta_{iter}$ is the attack step size at each iteration, and $N$ is the total iteration number. For each step, intuitively the latent vector $z$, which represents the general latent representation, is perturbed by $\lambda_k$ and guided by $\va_k$, the specific semantic latent direction, such as attributes "age" and "smile". Therefore, combining FTM and AGE, we propose the AGE-FTM, which is a dual manifold attack consisting of both off- and on-manifold attacks,
\begin{equation} \label{eq:AGE-FTM}
    {\boxed{\begin{split}
        & \text{\bf AGE-FTM optimization:} \\
        &\min_{\lambda \in \Lambda,~\delta \in \Delta} \gD (f_{\theta}(G(E(x) + \lambda; \va) + \delta \odot M, f_{\theta}(x_t)).
    \end{split}}}
\end{equation}

\section{Experiments}

\subsection{Settings} \label{sec: Settings}
\paragraph{Datasets} To generate high quality face images , the FFHQ dataset \cite{karras2019style} is used for training following \cite{wang2021HFGI}. To train the face parsing algorithm, the CelebAMask-HQ dataset \cite{CelebAMask-HQ} is used. CelebAMask-HQ is a large-scale face image dataset that has 30000 high-resolution face images, whose masks are manually-annotated including all facial components and accessories such as skin, nose and eyes. For testing, we choose CelebA-HQ dataset \cite{karras2017progressive} and FFHQ dataset \cite{karras2019style}, which are popular face image datasets with high resolution. For CelebA-HQ, we use a subset of 1000 face images with different identities following \cite{hu2022protecting}. For FFHQ, we randomly select $1000$ images.


\paragraph{Evaluation details} We compare the performance of the proposed method with different adversarial attacks specifically for FR: 1) Perturbation-based attack TIP-IM \cite{yang2021towards}; 2) Makeup-based attack Adv-Makeup \cite{yin2021adv}; and 3) Style transfer attack AMT-GAN \cite{hu2022protecting}. To evaluate the attack ability of different impersonation attacks, we follow \cite{xiao2021improving} to use attack success rate (ASR). We show the ASR at FAR@0.01 in a black-box setting. To evaluate the image quality of adversarial images, we use FID \cite{heusel2017gans}, which measures the distance of the data distribution. 
\paragraph{Implementation details} For texture extraction, the blurring operator $\gB$ is a Gaussian smoothing operator with kernal size $19 \times 19$ and the standard deviation is $5$. The threshold for texture extraction $\gamma$ is $0.003$. For face parsing, we used a pretrained  BiSeNet \cite{yu2018bisenet} trained on the CelebAMask-HQ dataset \cite{CelebAMask-HQ}. For the generator $G$, we used a pre-trained HFGI \cite{wang2021HFGI} trained on the FFHQ dataset \cite{karras2019style}. To obtain the attribute vectors, we apply InterfaceGAN \cite{shen2020interpreting} to obtain the attribute vectors to edit the generated images. 

We set $\epsilon=16/255$, $\epsilon_{iter}=2/255$ with fifty iteration steps for the off-manifold attack FTM. We set $\eta=0.1$, $\eta_{iter}=0.02$ with $N=10$ iteration steps for the on-manifold attack AGE. The pretrained FR models that we use for black box attacks includes IR152 \cite{he2016deep}, IRSE50 \cite{hu2018squeeze}, Facenet \cite{schroff2015facenet}, and Mobileface \cite{deng2019arcface}. All the experiments are conducted on one NVIDIA RTX 2080Ti GPU.

\begin{table}
\setlength\tabcolsep{1pt}
\centering
\small
\scalebox{0.8}{\begin{tabular}{@{}c|cccc|c||cccc|c@{}}
\toprule
 & \multicolumn{4}{c|}{Black-box Attack Ability} & {Image Quality} & \multicolumn{4}{c|}{Black-box Attack Ability} & {Image Quality}\\
 \midrule
                &  IRSE50  &  IR152  & Facenet  & Mobileface & FID ($\downarrow$)  &  IRSE50  &  IR152  & Facenet  & Mobileface & FID ($\downarrow$) \\
\midrule
Clean &7.29 & 3.80 & 1.08 &12.68  & -&4.4 & 2.5 & 1.7 &5.2  & -\\
\midrule
TIP-IM \cite{yang2021towards} & 54.4  &\underline{37.2}   &\textbf{40.7}   &48.7 & 38.74 & \textbf{52.3}&\underline{27.5} &\textbf{12.50} &\textbf{55.1} &39.86 \\
AMT-GAN \cite{hu2022protecting} & \textbf{77.0}  &35.1   &\underline{16.6}  &50.7 &34.44 &19.9 & 16.1& 3.7& 30.4&35.23  \\

\midrule
AGE-TMA (ours) & \underline{69.1} & \textbf{51.0} &13.2 &\textbf{78.0} &\textbf{27.61} &\underline{47.7} &\textbf{36.6} &\underline{11.2} &\underline{53.9} &\underline{32.04} \\
AGE-FTM (ours)&51.0 &34.7 &13.1 &\underline{63.6} & \underline{27.63} &28.0 & 27.4& 11.0&34.0 & \textbf{31.51}\\
\bottomrule
\end{tabular}}
\caption{Quantitative evaluation of the proposed method and other compared methods. Left: CelebA-HQ. Right: FFHQ.}
\label{table:Main}

\end{table}

\begin{figure*}[t]
\centering
\setlength\tabcolsep{1pt}
\scalebox{1.0}{\begin{tabular}{cccc}

\begin{subfigure}[t]{0.24\textwidth}
\includegraphics[width=\textwidth]{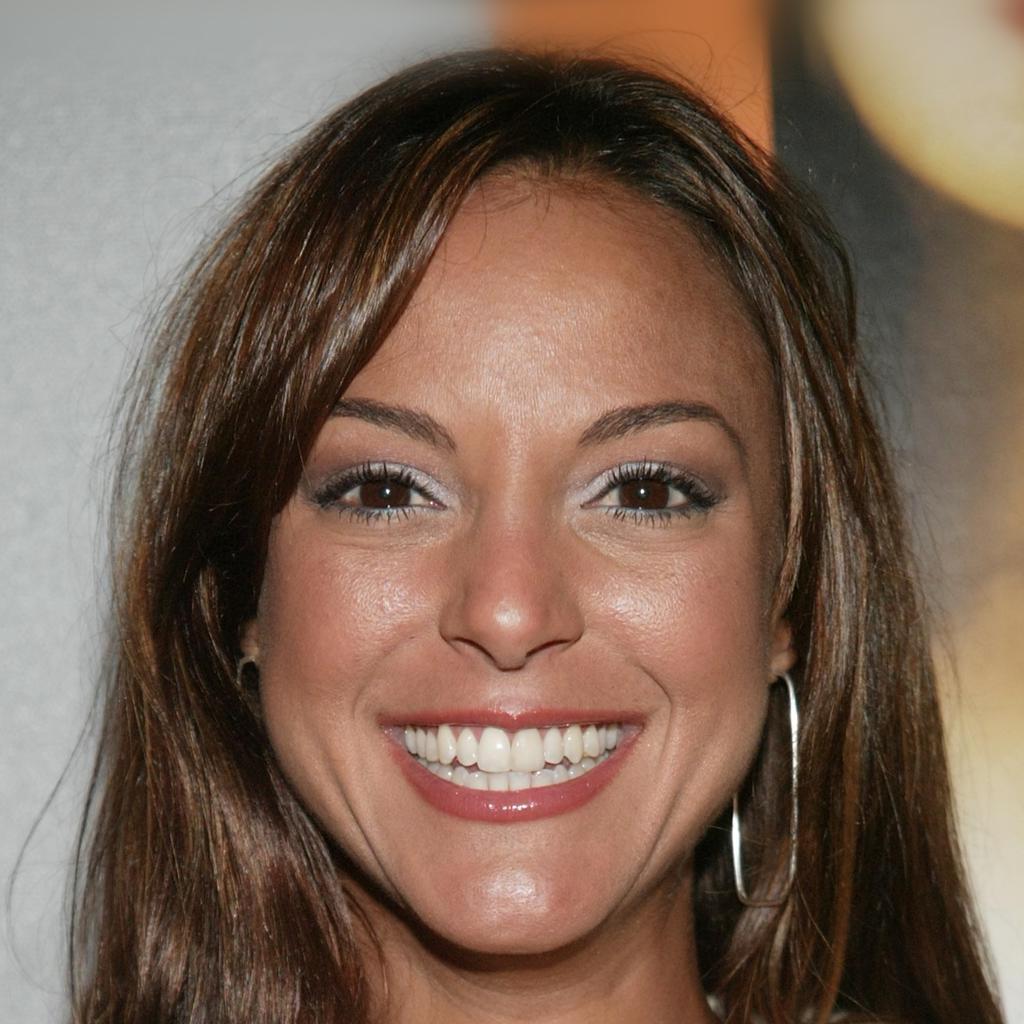}
\vspace{-11mm}
\begin{flushleft}
\small{{\textcolor{white}{\textbf{44.79}}}}
\end{flushleft}
\vspace{-2mm}
\end{subfigure} &
\begin{subfigure}[t]{0.24\textwidth}
\includegraphics[width=\textwidth]{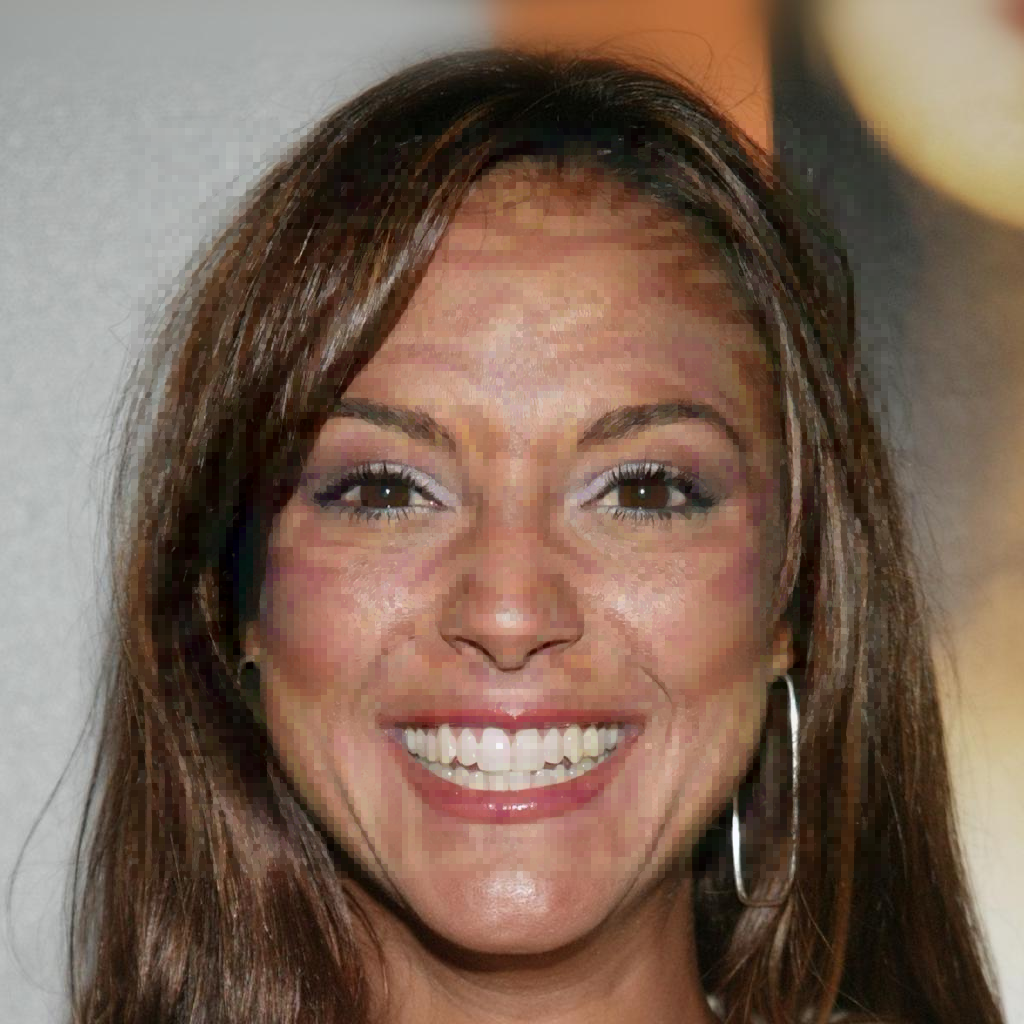}
\vspace{-11mm}
\begin{flushleft}
\small{{\textcolor{white}{\textbf{68.41}}}}
\end{flushleft}
\vspace{-2mm}

\end{subfigure} &
\begin{subfigure}[t]{0.24\textwidth}
\includegraphics[width=\textwidth]{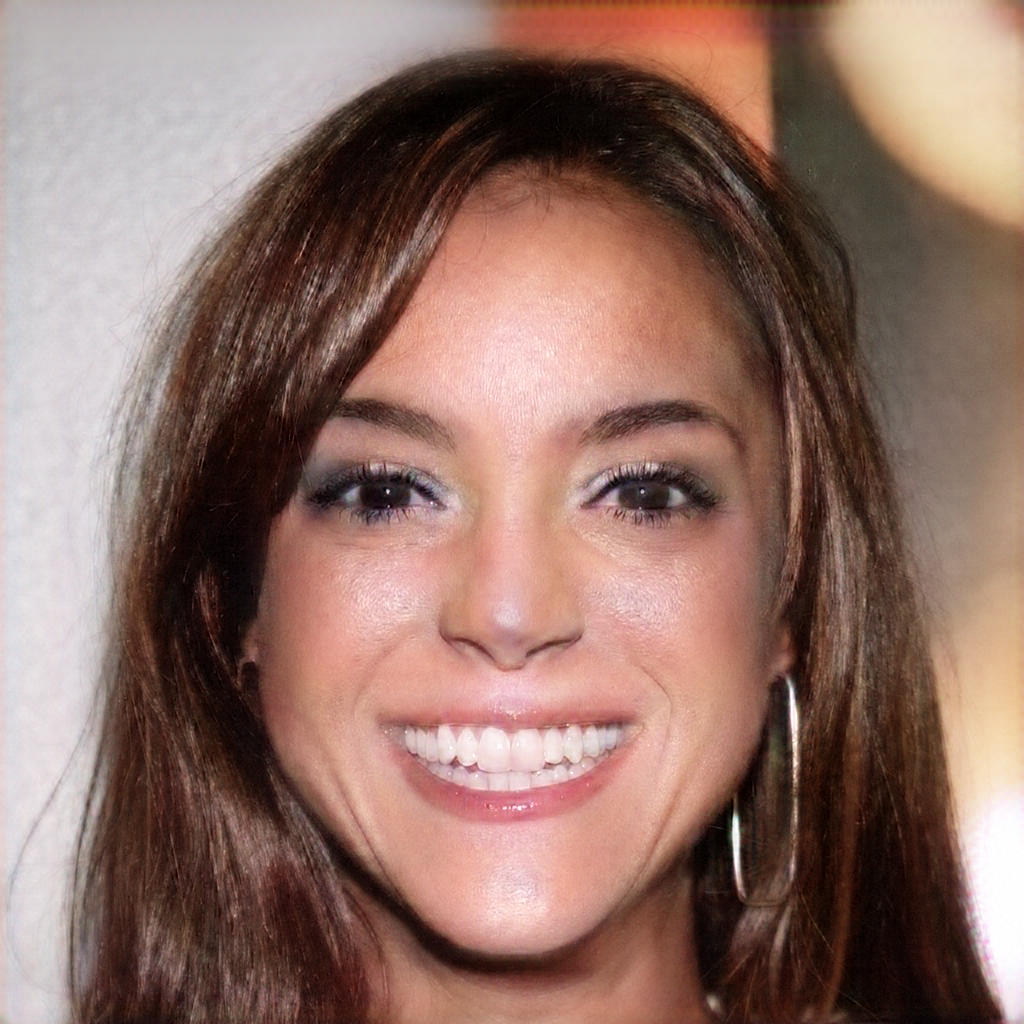}
\vspace{-11mm}
\begin{flushleft}
\small{{\textcolor{white}{\textbf{66.43}}}}
\end{flushleft}
\vspace{-2mm}

\end{subfigure} &
\begin{subfigure}[t]{0.24\textwidth}
\includegraphics[width=\textwidth]{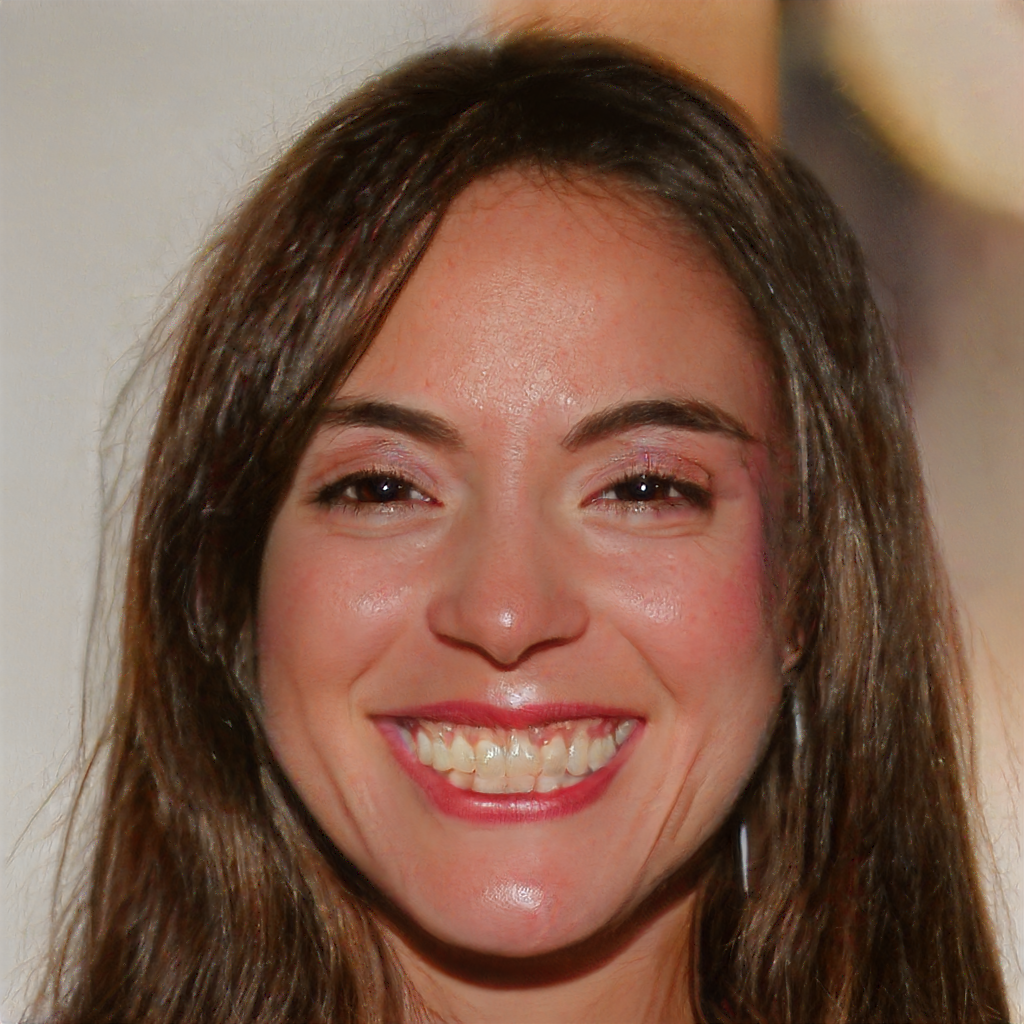}
\vspace{-11mm}
\begin{flushleft}
\small{{\textcolor{white}{\textbf{68.71}}}}
\end{flushleft}
\vspace{-2mm}

\end{subfigure} 
\\
\begin{subfigure}[t]{0.24\textwidth}
\includegraphics[width=\textwidth]{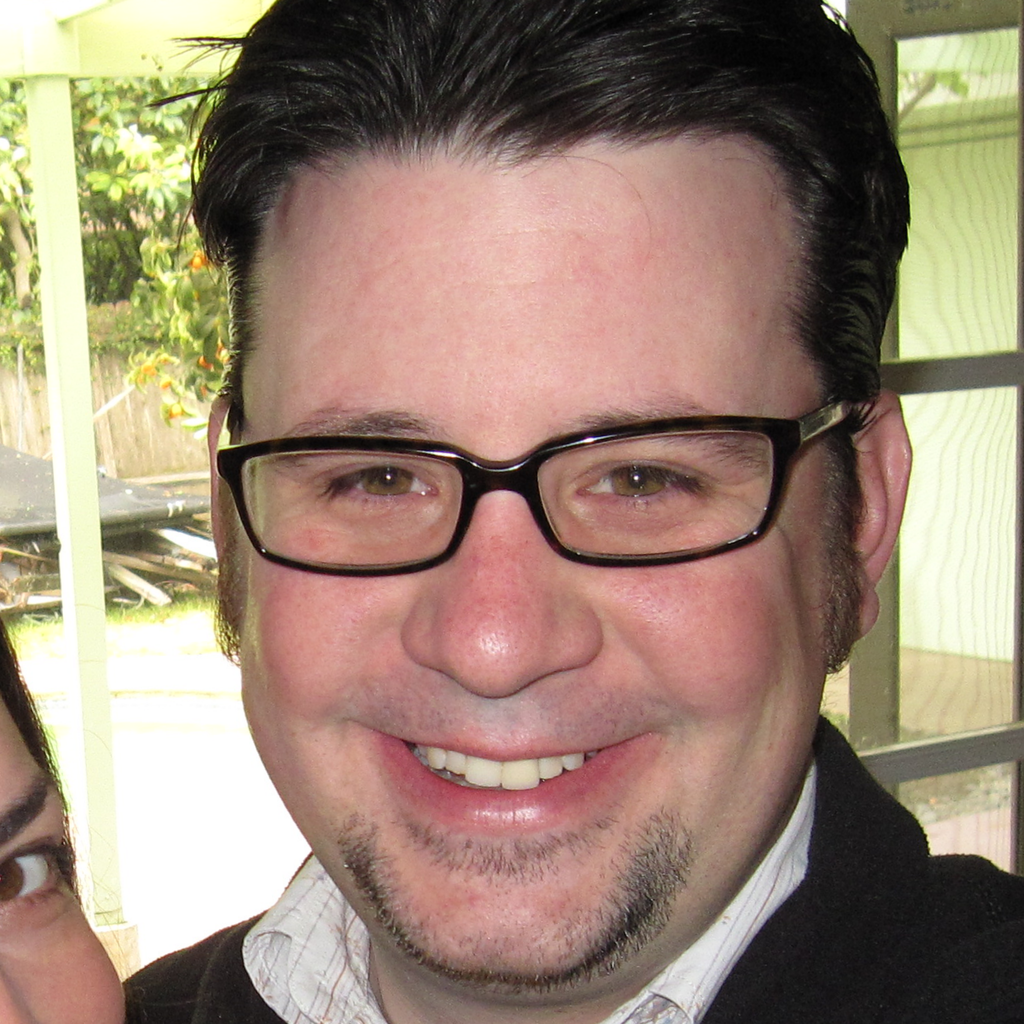}
\vspace{-11mm}
\begin{flushleft}
\small{\textcolor{white}{\textbf{20.38}}}
\end{flushleft}
\vspace{-2mm}
\caption{Original}
\end{subfigure} 
&
\begin{subfigure}[t]{0.24\textwidth}
\includegraphics[width=\textwidth]{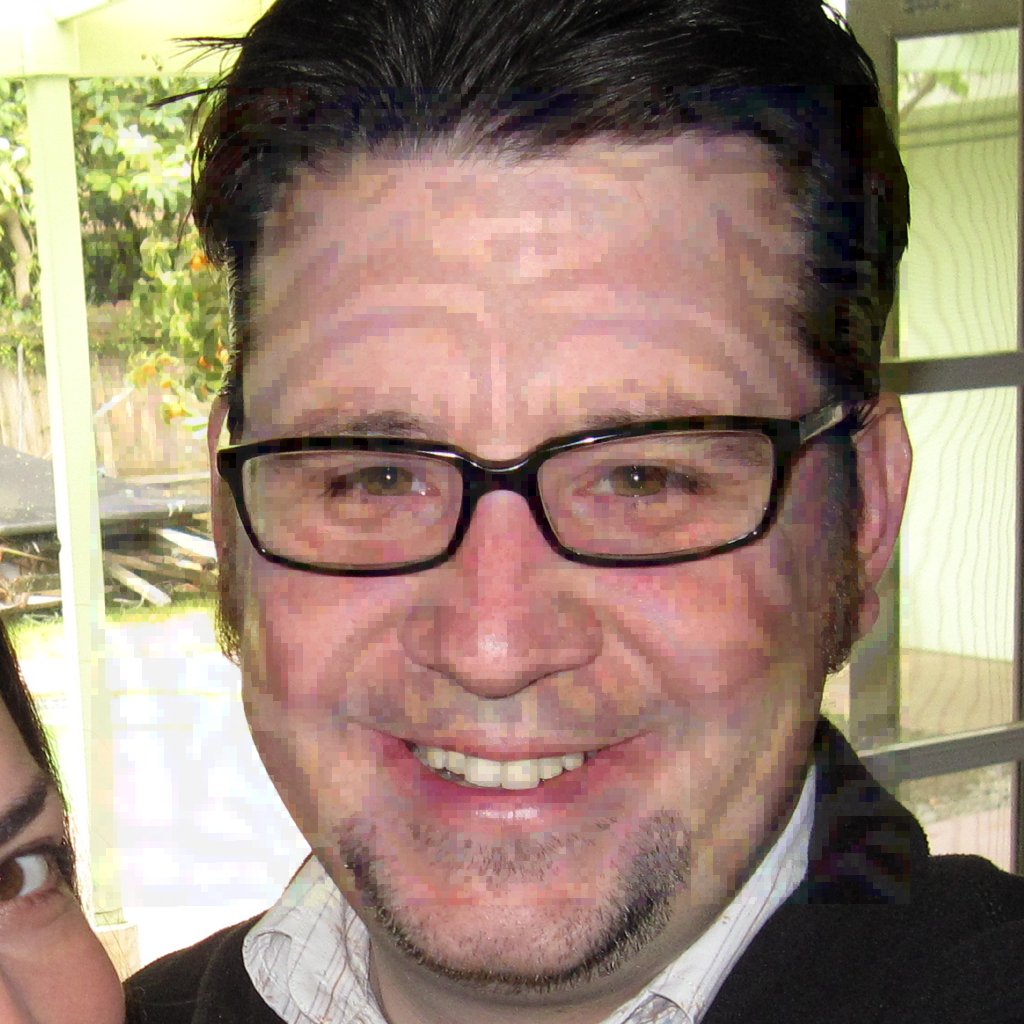}
\vspace{-11mm}
\begin{flushleft}
\small{\textcolor{white}{\textbf{45.87}}}
\end{flushleft}
\vspace{-2mm}
\caption{TIP-IM \cite{yang2021towards}}
\end{subfigure} &
\begin{subfigure}[t]{0.24\textwidth}
\includegraphics[width=\textwidth]{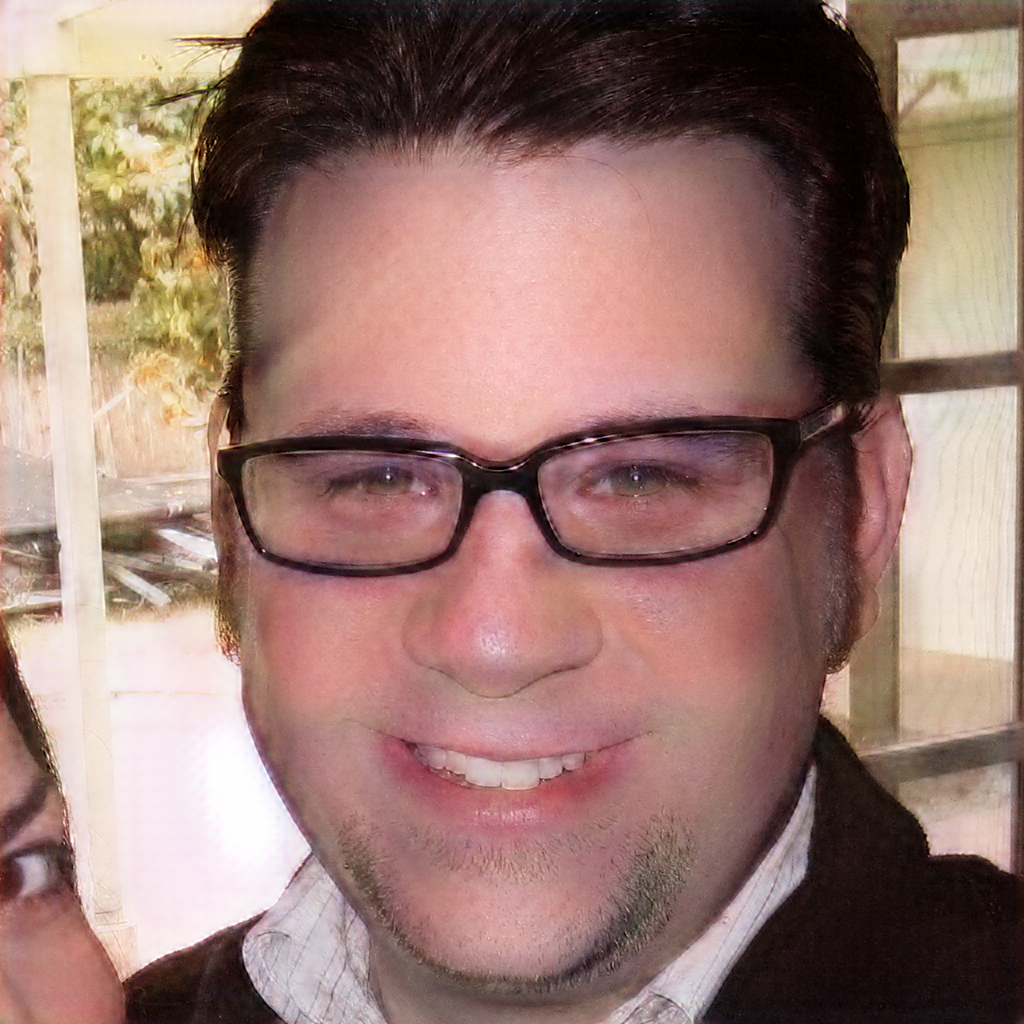}
\vspace{-11mm}
\begin{flushleft}
\small{\textcolor{white}{\textbf{28.96}}}
\end{flushleft}
\vspace{-2mm}
\caption{AMT-GAN \cite{hu2022protecting}}
\end{subfigure} &
\begin{subfigure}[t]{0.24\textwidth}
\includegraphics[width=\textwidth]{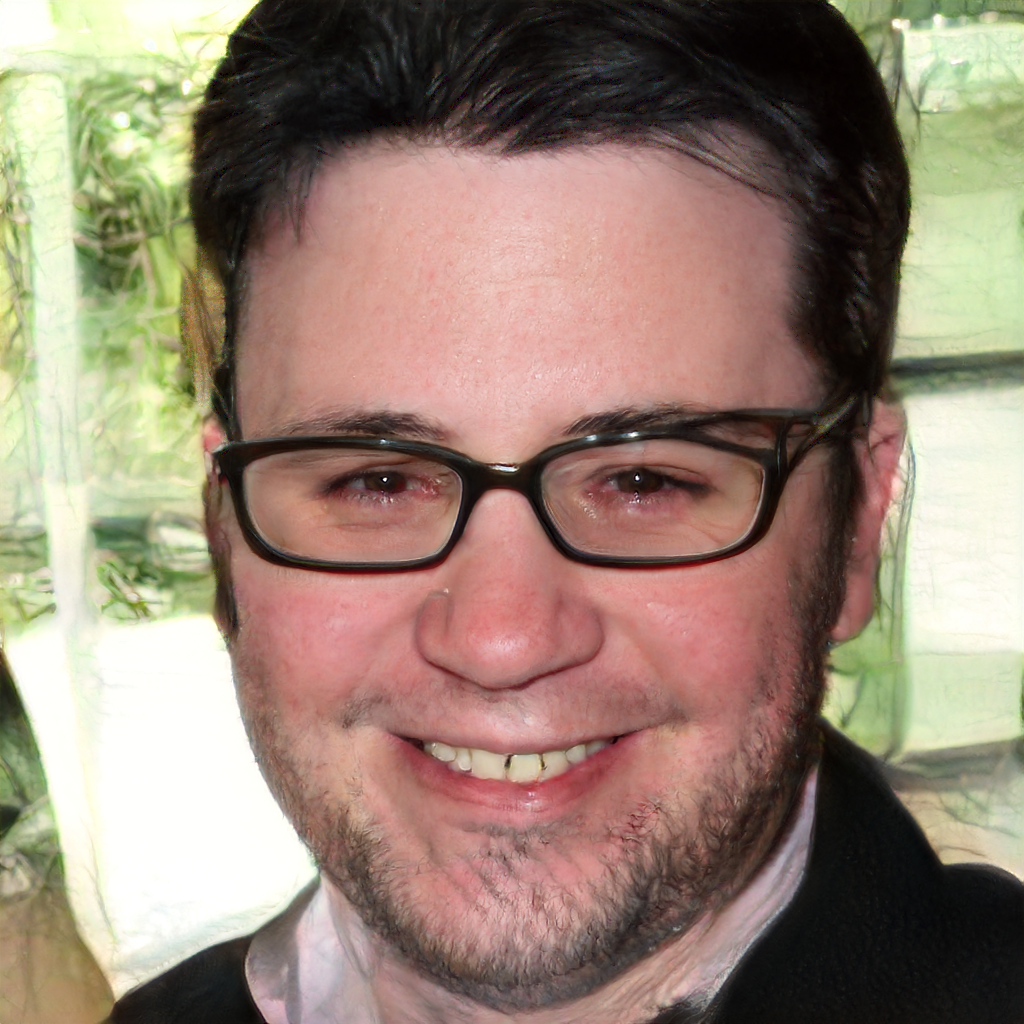}
\vspace{-11mm}
\begin{flushleft}
\small{\textcolor{white}{\textbf{30.12}}}
\end{flushleft}
\vspace{-2mm}
\caption{AGE-FTM (ours)}
\end{subfigure} 
\end{tabular}}

\caption{{Visualization of the proposed method AGE-FTM attack with other compared methods. Top: CelebA. Bottom: FFHQ.}}
\label{fig: compare}
\end{figure*}

\subsection{Main Results}

\paragraph{Black-box attack ability} The black-box attack results on four different pre-trained FR models are shown in Table \ref{table:Main}. For each target model, the other three FR models are used to craft the attacks. From Table \ref{table:Main}, we observe both AGE-TMA and AGE-FTM have a strong black-box attack ability. AGE-TMA has the strongest attack ability among all the methods in the CelebAMask-HQ dataset and is comparable TIP-IM in the FFHQ dataset. In both datasets, the proposed method is significantly better than AMT-GAN. 

\paragraph{Image quality}
The image quality results evaluated with FID are shown in Table \ref{table:Main}. To measure the naturalness of the attacked images, we also show the FID results. Note that the proposed methods have the best FID results. This demonstrates that the adversarial images generated by AGE-FTM have a more natural appearance, which is the goal of this paper. The visualization is in Fig. \ref{fig: compare}. We can observe similar results from FTM in Table \ref{table:Ablation studies}. Since FTM only attacks the textures on the hair region, it is a local attack. All the pixels which are not on the hair region are not perturbed. Therefore, it achieves good image quality results.
\begin{table}[t!]
\setlength\tabcolsep{2pt}

\centering
\small
\scalebox{1}{\begin{tabular}{@{}c|c|cc|ccc}
\toprule
Method        &     $\text{Clean}$   &    $\text{TMA}$  & $\text{FTM}$  & AGE  &  $\text{AGE-TMA}$   &  $\text{AGE-FTM}$ \\
\midrule
FID ($\downarrow$)  & - & 1.33  & 0.19  &27.64 &27.61 &27.63\\
ASR ($\uparrow$) & 12.68 & 38.9 &18.1 &53.0 &78.0 & 63.6 \\
\bottomrule
\end{tabular}}
\caption{Ablation studies of AGE-FTM. PSNR, SSIM, FID and ASR are evaluated with CelebA-HQ datase. ASR at FAR@$0.01$ is evaluated in a black-box setting with the targeted model Mobileface.} \label{table:Ablation studies}
\end{table}

\begin{figure*}[t]
\setlength\tabcolsep{1pt}
\centering
\scalebox{1.0}{\begin{tabular}{ccccc}

\begin{subfigure}[t]{0.19\textwidth}
\includegraphics[width=\textwidth]{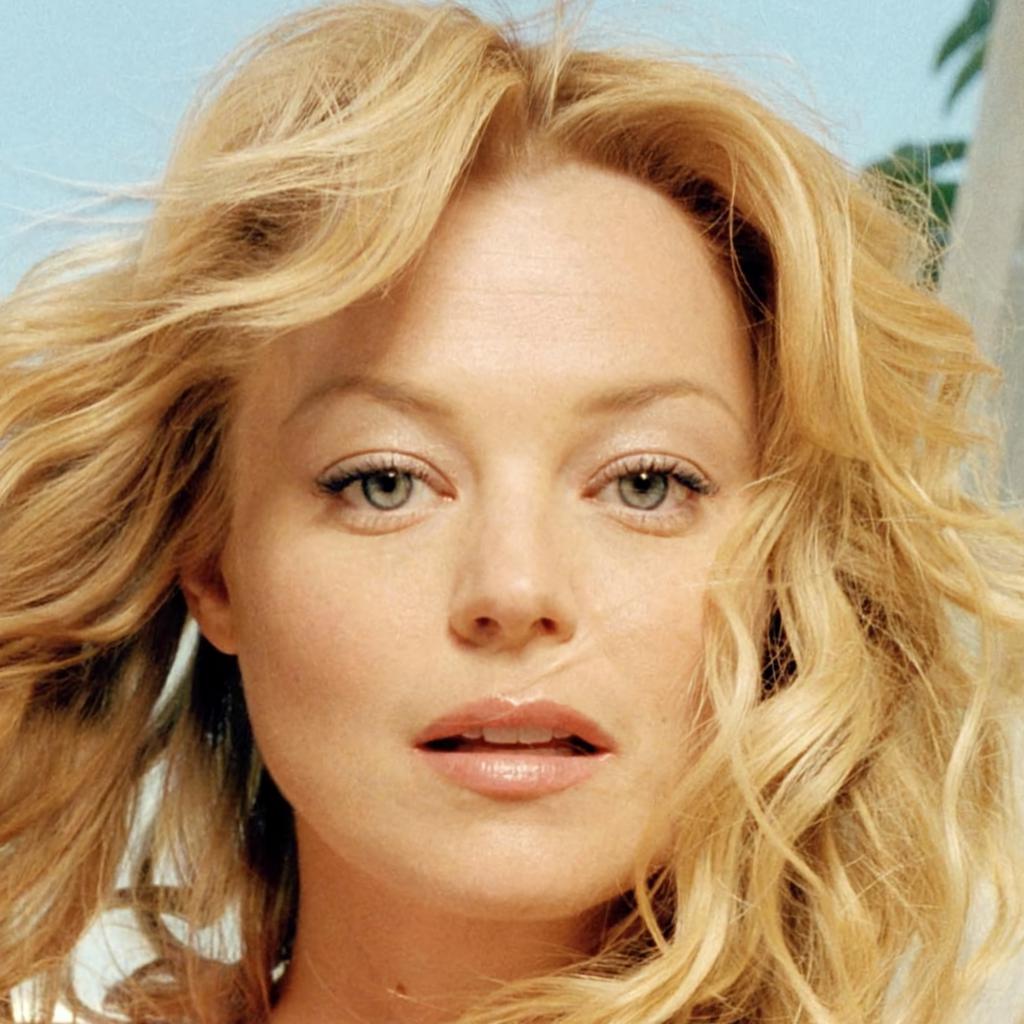}
\caption{Original}
\end{subfigure} &
\begin{subfigure}[t]{0.19\textwidth}
\includegraphics[width=\textwidth]{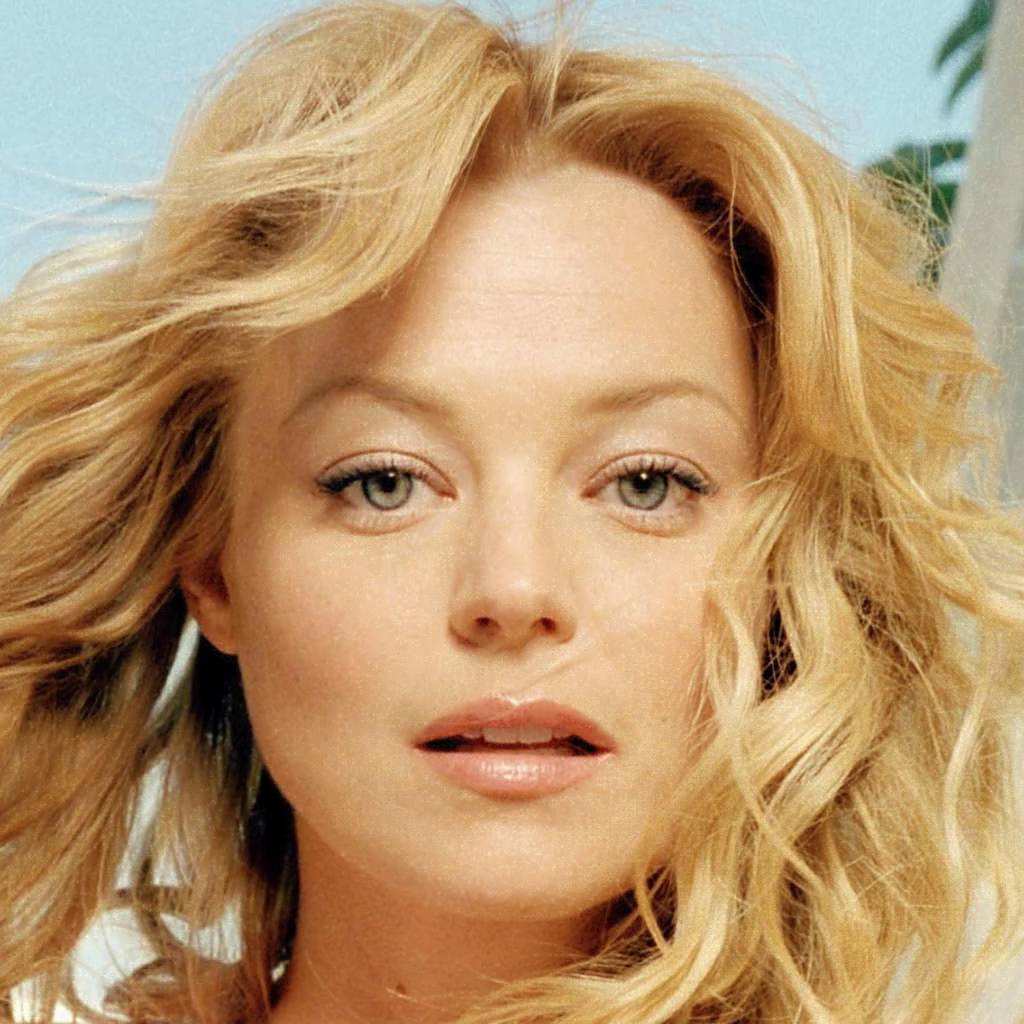}
\caption{TMA}
\end{subfigure} &
\begin{subfigure}[t]{0.19\textwidth}
\includegraphics[width=\textwidth]{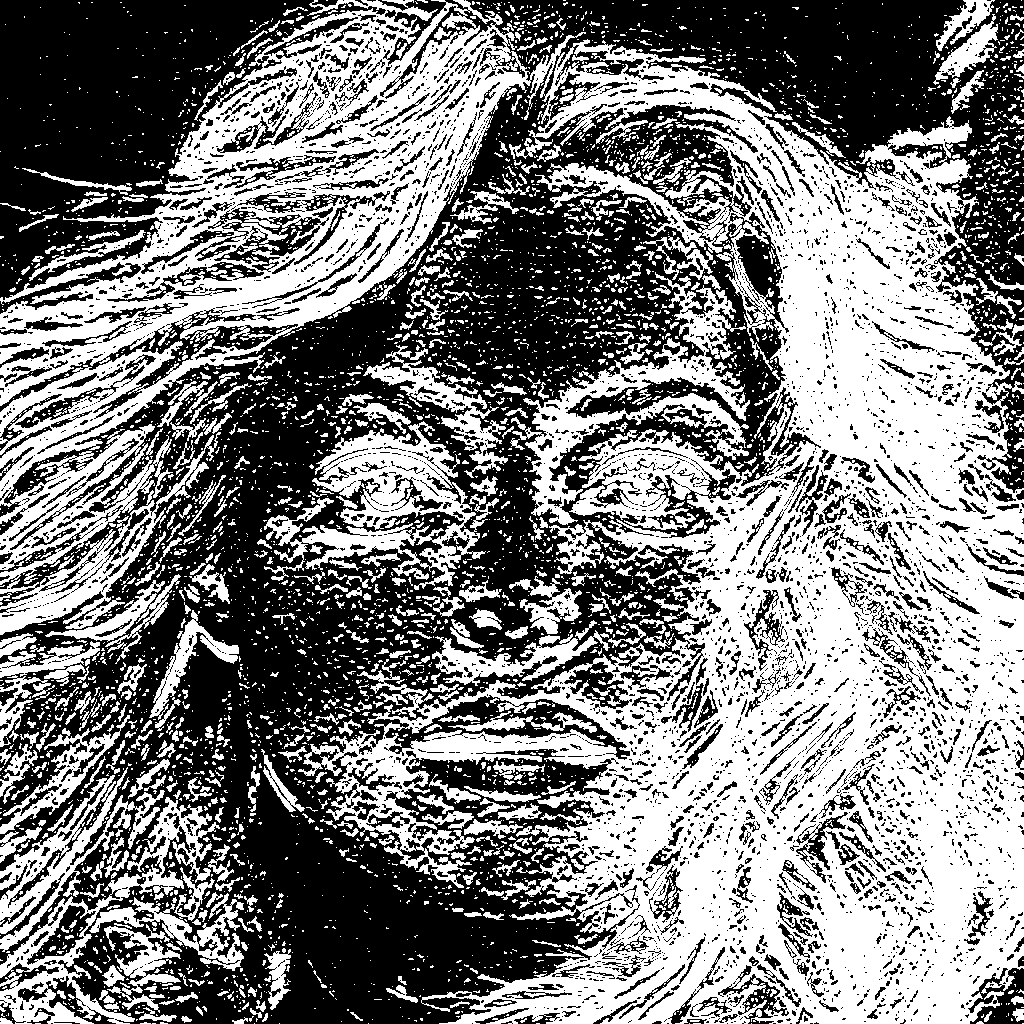}
\caption{TMA Mask}
\end{subfigure} &
\begin{subfigure}[t]{0.19\textwidth}
\includegraphics[width=\textwidth]{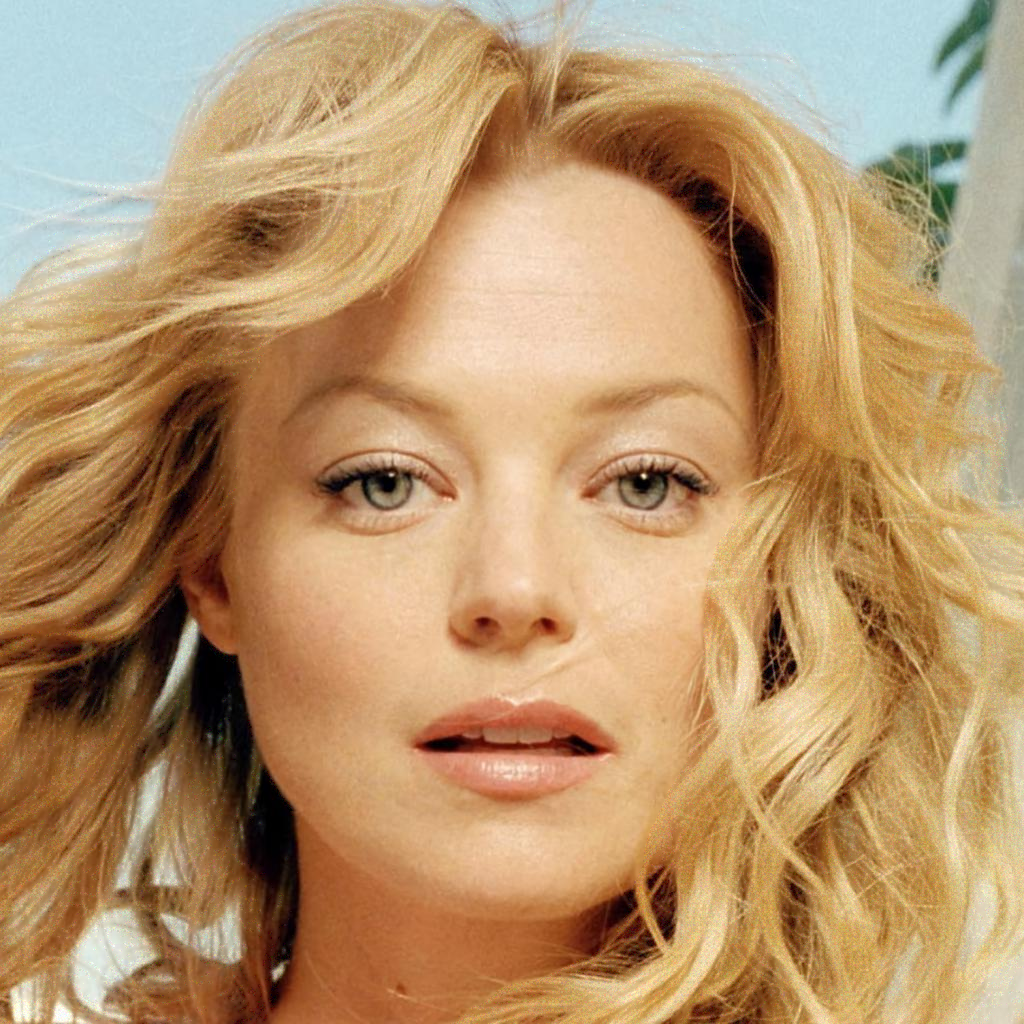}
\caption{FTM}
\end{subfigure} &
\begin{subfigure}[t]{0.19\textwidth}
\includegraphics[width=\textwidth]{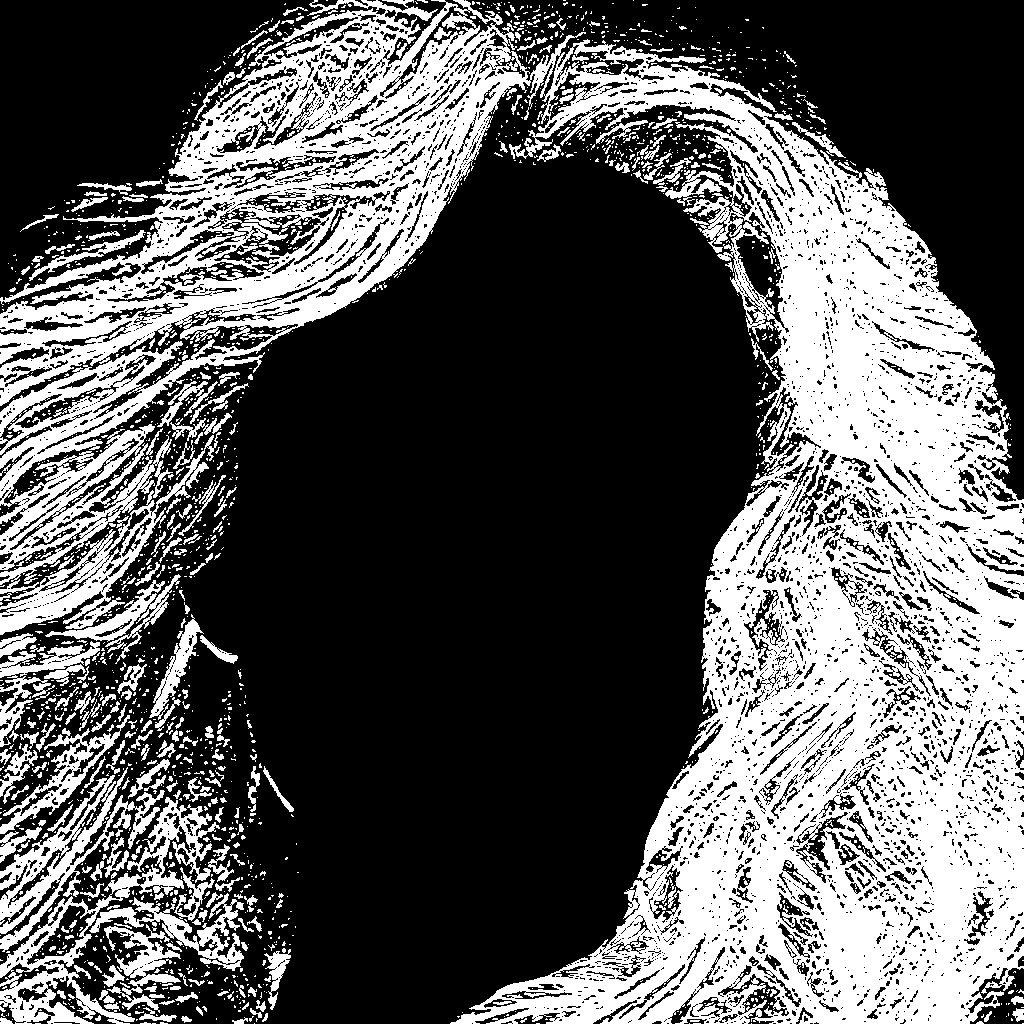}
\caption{FTM mask}
\end{subfigure} 
\end{tabular}}

\caption{Visualization of the proposed FTM attack. For TMA, it is harder to notice as we only perturb the texture of the image. To further improve the imperceptibility, FTM only perturbs the texture of the hair portion of the face image. This makes it even harder to distinguish whether it is attacked or not.}
\label{fig: FTM demo}
\end{figure*}

\begin{figure*}[t]
\setlength\tabcolsep{1pt}
\centering
\scalebox{1.0}{\begin{tabular}{cccc}
\begin{subfigure}[t]{0.24\textwidth}
\includegraphics[width=\textwidth]{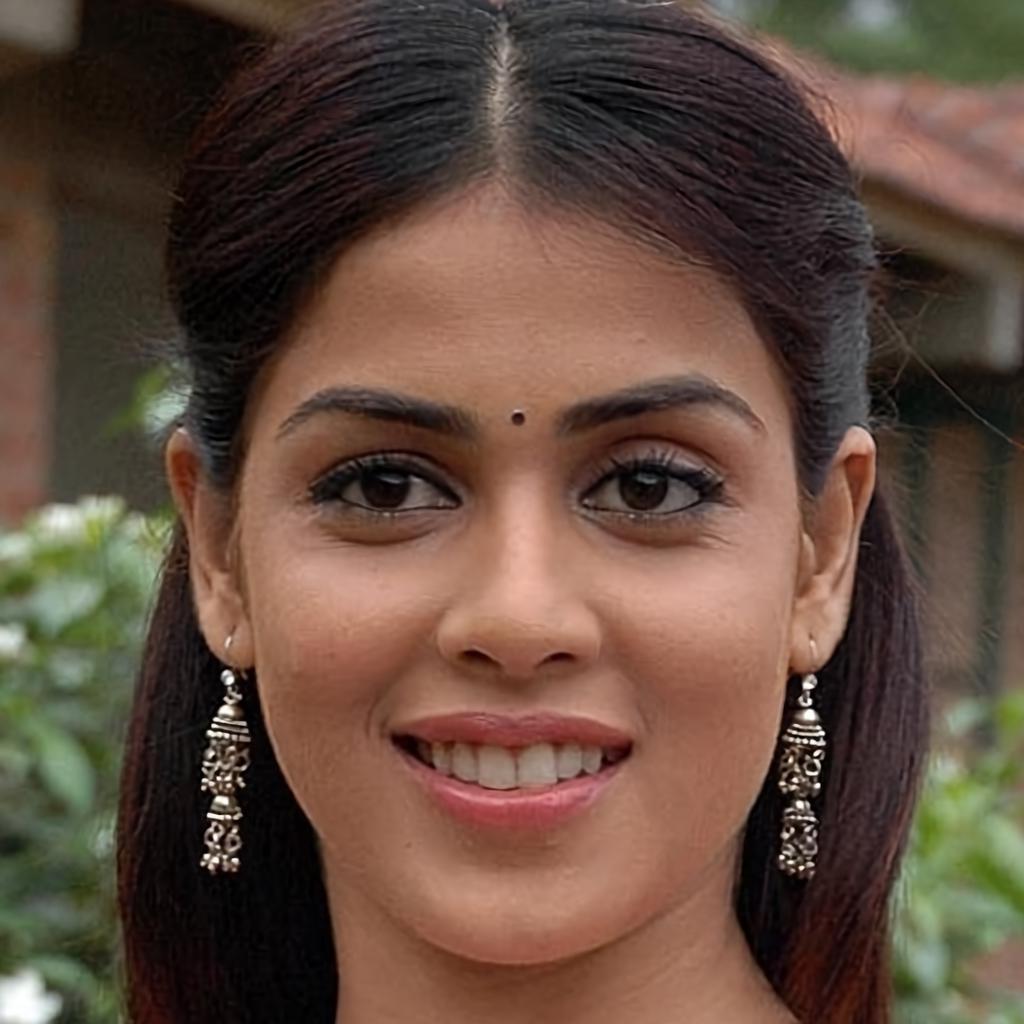}
\caption{Original}
\end{subfigure} &
\begin{subfigure}[t]{0.24\textwidth}
\includegraphics[width=\textwidth]{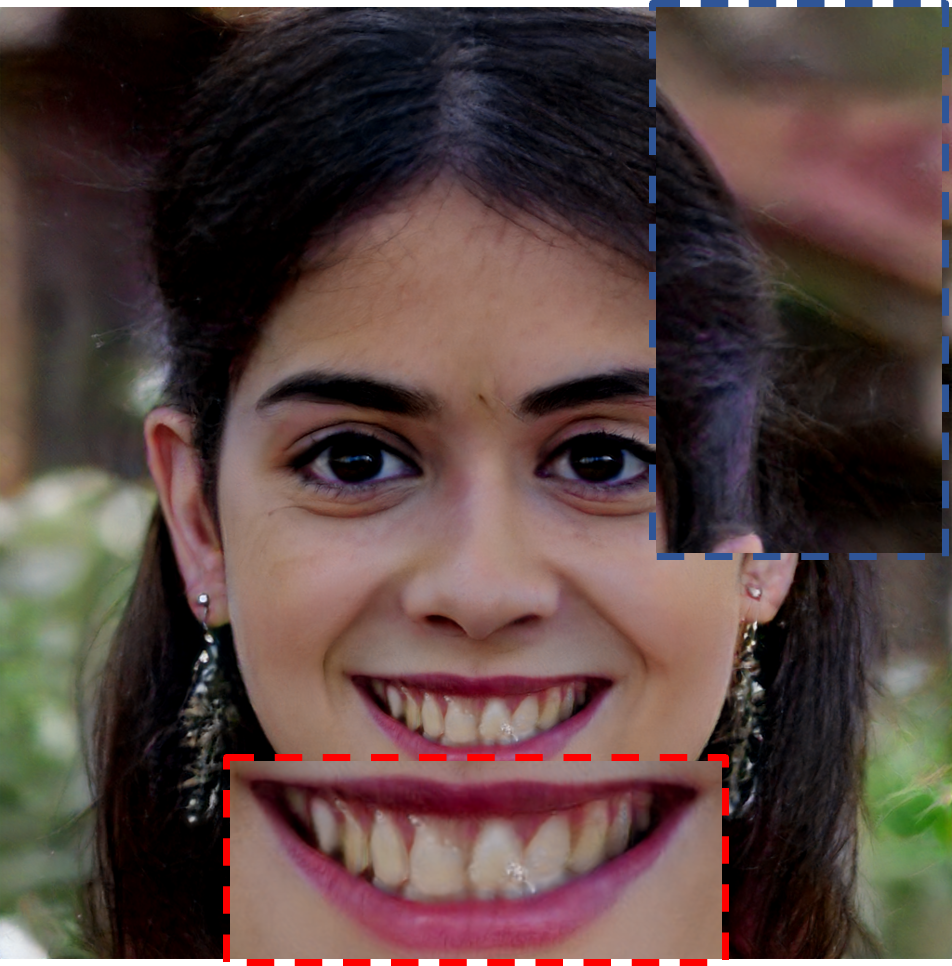}
\caption{AGE}
\end{subfigure} &
\begin{subfigure}[t]{0.24\textwidth}
\includegraphics[width=\textwidth]{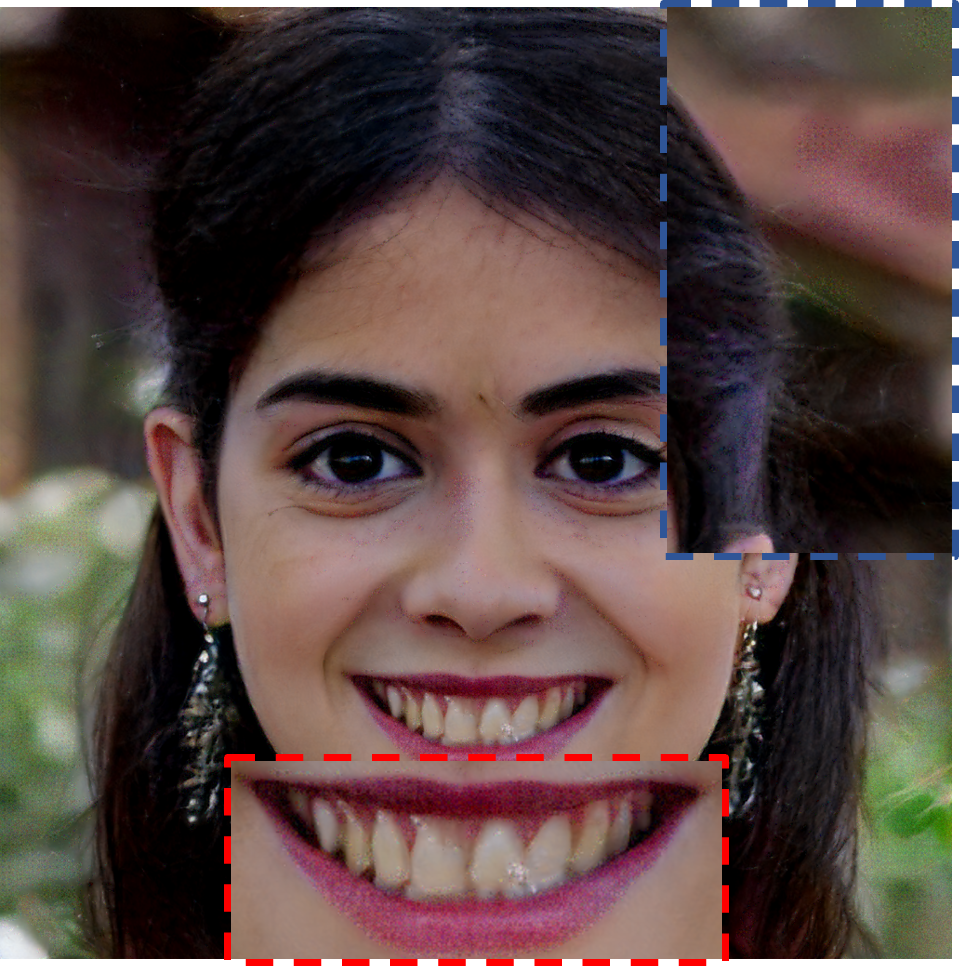}
\caption{AGE-TMA}
\end{subfigure} &
\begin{subfigure}[t]{0.24\textwidth}
\includegraphics[width=\textwidth]{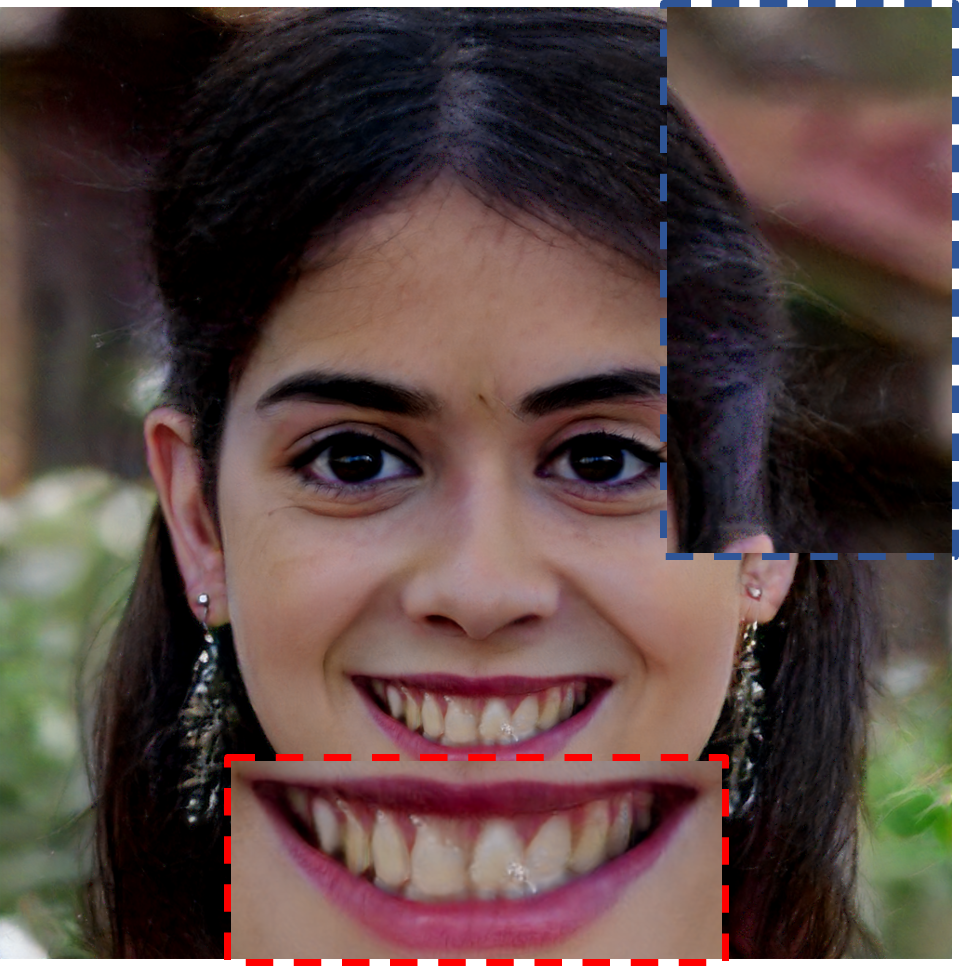}
\caption{AGE-FTM}
\end{subfigure} 
\end{tabular}}

\caption{Visualization of the proposed FTM attack. For PGD attack, all the pixels are perturbed and it is easy to notice the perturbation. For TMA, it is harder to notice as we only perturb the texture of the image. To further improve the imperceptibility, FTM only perturbs the texture of the hair portion of the face image. This makes it even harder to distinguish whether it is attacked or not.}
\label{fig: AGE demo}
\end{figure*}



\subsection{Ablation Studies}

We show the ablation studies evaluating the image quality and the attack ability of the proposed method in Table \ref{table:Ablation studies}. The ASR is evaluated at FAR@$0.01$ with Mobileface as the targeted model. We trained five models for evaluating the performance of the proposed methods AGE-FTM. 

\paragraph{FTM} To compare the facial masking component, we have 1) the TMA that only attacks the textures of the image; and 2) the FTM attack that only attacks the textures of the hair portion. The visualization is shown in Fig. \ref{fig: FTM demo}. From Table \ref{table:Ablation studies}, we can observe FTM achieves the best image quality. On the other hand, we can see there is a trade-off between the image quality and the attack ability. This is reasonable as the total number of perturbed pixels is reduced significantly for the case of TMA and FTM (See Fig. \ref{fig: FTM demo}(d) and (f)). In particular for FTM, since the attack is restricted to the hair portion, the attack is weakened. 


\paragraph{AGE} To compare the AGE component, we have 1) AGE which uses the attribute vector to guide the on-manifold attack; 2) AGE-TMA that is a dual manifold attack containing AGE and TMA; and 3) AGE-FTM that is our ultimate dual manifold attack containing AGE and FTM. The visualization is shown in Fig. \ref{fig: AGE demo}. From Table \ref{table:Ablation studies}, we can see that AGE itself achieves descent ASR. Similar to makeup attacks, AGE also changes the style/attribute of the images. However, the proposed AGE is different from other makeup attacks as the attribute vectors, which codes the semantic latent vector directions, are used to regularize the on-manifold perturbation. The noise added in TMA and FTM is hardly noticeable (See Fig. \ref{fig: AGE demo}(c) and (d)).

\subsection{Attack Ability on Commercial FR API}
We evaluate the attack performance by collecting and averaging the confidence score from the commercial FR API Face++ with the CelebA-HQ dataset. The average confidence score for the clean images, TIP-IM~\cite{yang2021towards}, AMT-GAN~\cite{hu2022protecting}, and the proposed method AGE-FTM are 33.67\%, 57.06\%, 54.17\%, and 54.96\% respectively. This shows that the proposed method has a satisfactory attack ability even for the commercial API while achieving good image quality. 

\subsection{User Study}
We conducted a user study to evaluate the subjective image quality of our proposed method and baselines. The participants include forty five social media users. We showed the participants encrypted images of ten subjects from the CelebA-HQ dataset generated by TIP-IM~\cite{yang2021towards}, AMT-GAN~\cite{hu2022protecting}, and AGE-FTM (ours), and the participants were asked to choose up to three images for each person that they will like to use for sharing on social media if they were the person. On average, 68.89\% of the participants chose the image generated by AGE-FTM, 33.56\% chose AMT-GAN, and 6.22\% chose TIP-IM. The results demonstrate that our proposed method has higher image quality than the baselines and is more likely to be used in practice. 

\subsection{Limitations and Future direction}
AGE-FTM involves two components, AGE and FTM. AGE perturbs the latent space guided with attribute vectors to perform an on-manifold attack. However, it requires a generator that has a semantic latent space and generates a high-quality face image. A good face alignment is necessary for these face image GAN models. In other words, AGE does not work without face alignment. This would be a future direction that uses a different generator, such as flow-based or diffusion models that do not require face alignment.

\section{Conclusion}
In this paper, we propose AGE-FTM that performs dual manifold adversarial attack on FR systems for facial privacy protection. AGE-FTM utilizes the image manifold information learned by GAN models to create \textit{natural} changes to the source images guided by facial attributes, and adds \textit{imperceptible} perturbations through facial texture masking. Our extensive experiments on the CelebA-HQ dataset demonstrate that AGE-FTM achieves state-of-the-art black-box attack performance and good visual quality, which can be useful for facial privacy protection in real-world applications.

\bibliography{ref}
\end{document}